\documentclass{article}
\usepackage{ijcai19}

\usepackage{times}
\usepackage{soul}
\usepackage{url}
\usepackage[hidelinks]{hyperref}
\usepackage[utf8]{inputenc}
\usepackage[small]{caption}
\usepackage{graphicx}
\usepackage{amsmath}
\usepackage{booktabs}
\usepackage{algorithm}
\usepackage{algorithmic}
\usepackage{latexsym}

\usepackage{multirow}
\usepackage{amssymb}
\usepackage[english]{babel}
\usepackage{hyphenat}
\makeatletter
\newcommand{\ssymbol}[1]{$^{\@fnsymbol{#1}}$}
\makeatother

\let\vec\boldsymbol

\newcommand{\specialcell}[2][c]{\begin{tabular}[#1]{@{}c@{}}#2\end{tabular}}

\title{Exploring and Distilling Cross-Modal Information for Image Captioning}

\author{
}

\author{
Fenglin Liu$^1$\thanks{Equal contributions.}\and
Xuancheng Ren$^2$\footnotemark[1]\and
Yuanxin Liu$^3$\and
Kai Lei$^4$\thanks{Corresponding authors.}\And
Xu Sun$^2$\footnotemark[2]\\
\affiliations
$^1$ADSPLAB, School of ECE, Peking University\\
$^2$MOE Key Laboratory of Computational Linguistics, School of EECS, Peking University\\
$^3$School of ICE, Beijing University of Posts and Telecommunications\\
$^4$ICNLAB, School of ECE, Peking University \\
\emails
fenglinliu98@pku.edu.cn,
renxc@pku.edu.cn,
yuanxinLIU@bupt.edu.cn \\
leik@pkusz.edu.cn,
xusun@pku.edu.cn
}

\begin{document}

\maketitle

\begin{abstract}
Recently, attention-based encoder-decoder models have been used extensively in image captioning. Yet there is still great difficulty for the current methods to achieve deep image understanding. In this work, we argue that such understanding requires visual attention to correlated image regions and semantic attention to coherent attributes of interest. Based on the Transformer, to perform effective attention, we explore image captioning from a cross-modal perspective and propose the Global-and-Local Information Exploring-and-Distilling approach that explores and distills the source information in vision and language. It globally provides the aspect vector, a spatial and relational representation of images based on caption contexts, through the extraction of salient region groupings and attribute collocations, and locally extracts the fine-grained regions and attributes in reference to the aspect vector for word selection. Our Transformer-based model achieves a CIDEr score of 129.3 in offline COCO evaluation with remarkable efficiency in terms of accuracy, speed, and parameter budget.

\end{abstract}

\section{Introduction}

Image captioning is a very challenging yet pragmatic multi-discipline task that combines image understanding and language generation. The deep neural networks, especially the models based on the encoder-decoder framework, have shown great success in pushing the state-of-the-art image captioning \cite{yao2018exploring}. A modern solution is to exploit a convolutional neural network (CNN), e.g., ResNet \cite{he2016deep}, to encode the image and a recurrent neural network (RNN), e.g., LSTM \cite{hochreiter1997long}, to generate the sentence with attention mechanisms \cite{xu2015show} extracting relevant information. Considerable efforts are put to improve the framework, such as incorporating object-oriented image representations \cite{anderson2018bottom} and augmenting the information source with predicted textual attributes \cite{fang2015captions}. Recently, several studies try to consider the problem from a cross-modal way, making use of both image regions and textual attributes \cite{yao2017boosting,jiang2018learning,liu2018simnet,liu2019MIA}.

However, there is still great difficulty in deep image understanding, as the systems tend to view the image as unrelated individual parts \cite{brendel2019bagnet,geirhos2019bias} and are not guided to comprehend the general correlations of such parts. 
For example, given the word \textit{umbrella}, a person would likely associate it with the notion of \textit{rain} or the act of \textit{holding}, which is generally not learned by the existing systems regardless of the image representations they use. In this work, we argue that such understanding requires effective attention to correlated image regions and coherent attributes of interest, so that the systems could learn generalized combinations through observations from images and written captions.

\begin{figure}[t]

\centering
\includegraphics[width=1\linewidth]{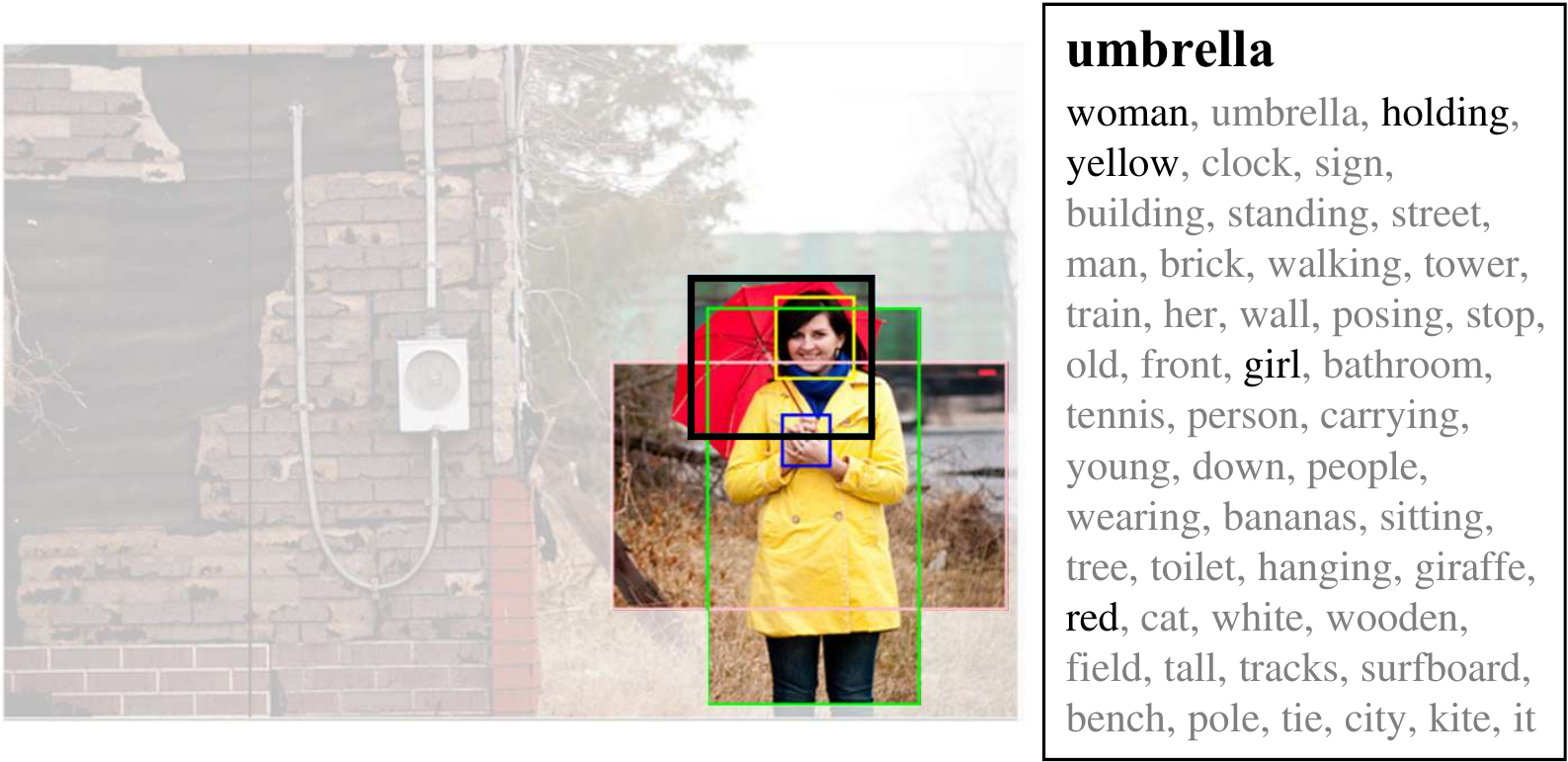}

\caption{Individual components of image representations often embody intrinsic combinations, which is beneficial for deep and semantic understanding of images. For visual regions, the focus on the umbrella is naturally extended to the related areas. For attribute words, the input word \textit{umbrella} is associated with common collocations. Those are learned by the proposed approach.}
\label{fig:motivation}
\end{figure}

\begin{figure}[t]

\centering
\includegraphics[width=1.0\linewidth]{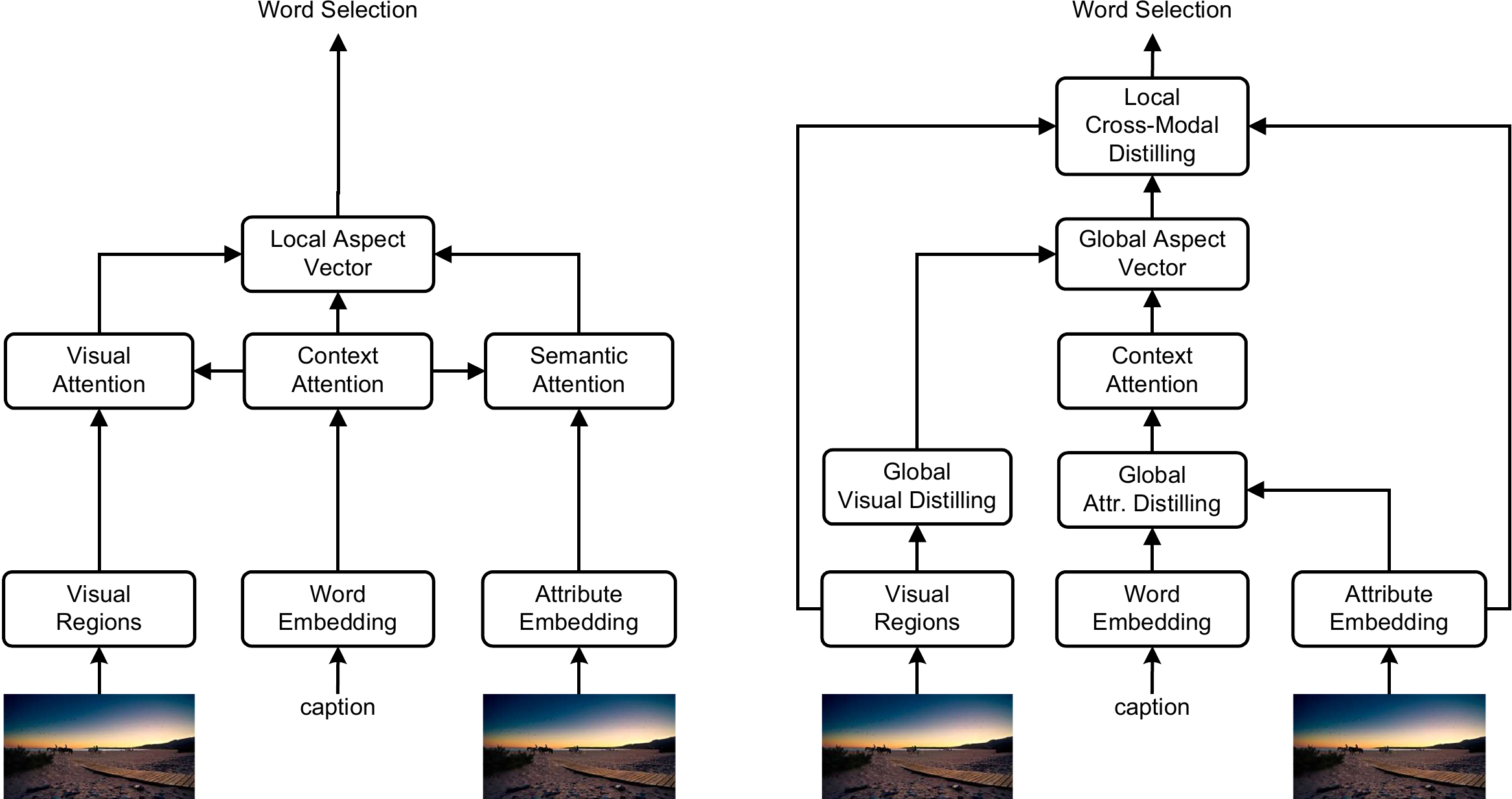}

\caption{Illustration of the difference between our cross-modal fully-attentive base model (Left) and the proposed model that distills the source information both globally and locally (Right).}
\label{fig:compare}
\end{figure}

In order to achieve that, we propose the Global-and-Local Information Exploring-and-Distilling approach that explores and distills the cross-modal source information. It first distills individual parts of image representations into inherently grouped image regions and attribute words, to form a spatial and relational representation. Those representations are not coupled with specific captions but contain general, informative, associative knowledge related to the image. Then, considering the current partially-generated captions, it provides globally the aspect vector, which semantically expresses and explores all the related representation groups that should be considered for the next output word. As the global aspect vector may not be rich in details, GLIED revisits individual parts of images and distills the regions and attributes again to form the local aspect vector, which could be more precise and finer-grained. 
We implement the approach upon a fully-attentive decoder using cross-modal representations. Sketches of both the base model and the proposed approach are shown in Figure~\ref{fig:compare}. The experiments on COCO image captioning dataset validate our argument and prove the effectiveness of the proposed approach.

Overall, the main contributions of this work are:
\begin{itemize}

\item We propose the Global-and-Local Information Exploring\hyp{}and\hyp{}Distilling approach, which  globally captures the inherent spatial and relational groupings of the individual image regions and attribute words for an aspect-based image representation, and locally it extracts fine-grained source information for precise and accurate word selection.

\item The experiments based on a fully-attentive decoder on the COCO image captioning dataset prove the effectiveness of our approach, which achieves 129.3 in terms of CIDEr with fewer parameters and faster computation, compared with existing state-of-the-art systems.

\item Further analysis shows that the proposed approach excels at generating complete descriptions and the learned region groupings and attribute collocations are in accordance with human intuition, which forms a powerful basis for describing images.

\end{itemize}

\section{Related Work}

Attention-based encoder-decoder models are used extensively in modern image captioning systems. Our work closely relates to the efforts on refining source image representations, using cross-modal information, and exploring semantic relationships for better image understanding.

\paragraph{Refining source representations.} To represent images, visual features extracted by CNNs are most-widely used \cite{xu2015show}, while textual features consisting of attribute word vectors are also proposed \cite{wu2016what}. Those kinds of features are often used by the decoder with the help of attention mechanisms to focus on the most relevant image regions or attribute words instead of the whole image, namely, visual attention \cite{xu2015show} or semantic attention \cite{you2016image}.  Visual features based on Region-CNNs and predicted bounding boxes \cite{anderson2018bottom} further extract object-oriented regions instead of generic regions considered by normal CNNs. Regardless of the type of source representations, relationships among the individual parts of representations (regions or attributes) are not defined, which should be essential to a semantic understanding of images.

\paragraph{Using cross-modal information.}
To our knowledge, there are some efforts \cite{jiang2018learning,yao2017boosting} trying to use both kinds of features in a non-trivial way. 
\citeauthor{jiang2018learning}~\shortcite{jiang2018learning} proposed to use a linear layer with max pooling to select guiding attributes as additional input for a recurrent decoder using visual attention.
\citeauthor{yao2017boosting}~\shortcite{yao2017boosting} presented a series of models (LSTM-A2,3,4,5) to combine visual features and textual attributes. 
Notably, those systems are based on recurrent decoders, while our approach is implemented on a fully-attentive decoder. \citeauthor{zhu2018captioning}~\shortcite{zhu2018captioning} also used a fully-attentive decoder but they did not consider cross-modal information and only incorporated visual attention.

\paragraph{Exploring semantic relationships.}   A new advance \cite{yao2018exploring} tried to explore visual relationships explicitly by using graph networks to encode scene graphs modeling the spatial and semantic relationships of image regions. 
However, the relationships between visual objects are predicted by a separate model with extra annotated data. In contrast, our model associates the features based on attention and the relationships are implicitly modeled as weighted combinations and trained together with the captioning model.

\section{Approach}

We first introduce the cross-modal information we used and our fully-attentive base decoder. Then, we present the global distilling method for extracting salient region groupings and attribute collocations, and the local distilling method for extracting fine-grained source information.

\subsection{Information Sources}

Since we consider the problem from a cross-modal point of view, we use both kinds of source representations, namely visual features and textual attributes. Visual features are good at illustrating the shapes and the colors, while it is not informative in what the item is in words. Textual attributes, in contrast, represent an image with high-level semantic concepts. Typically, the concepts are words that describe objects (e.g. \textit{person}, \textit{car}), attributes (e.g. \textit{off}, \textit{electric}), or relationships (e.g. \textit{using}, \textit{sitting}). 
For visual regions, we utilize the RCNN-based features, following the implementation of \citeauthor{anderson2018bottom}~\shortcite{anderson2018bottom}. We denote the extracted image regions as $I$. For textual attributes, we adopt Multiple Instance Learning \cite{zhang2006multiple}, a weakly-supervised method, to build an attribute extractor, following \citeauthor{fang2015captions}~\shortcite{fang2015captions}. 
Since textual attributes are discrete word tokens, we use an attribute embedding to project them into vectors, which are then stacked as visual attributes $A$.

\subsection{Cross-Modal Base Model}

Our cross-modal base model is adapted from \cite{ashish2017attention}, which is a neural model entirely driven by attention mechanisms without recurrent connections, and further incorporates the visual attention and the semantic attention specific to the image captioning task, making up a fully-attentive captioning decoder, which is shown in the left of Figure~\ref{fig:compare}.

The basic building block is the multi-head attention, and each head is defined as a scaled dot-product:
\begin{align}
\footnotesize
\mathcal{A}(Q,K,V)_i = \text{softmax}\left(\frac{{Q}{W}^{Q}_{i}({K}{W}^{K}_{i})^\mathsf{T})}{\sqrt{{d}_{k}}}\right){V}{W}^{V}_{i}
\end{align}
where  ${Q} \in \mathbb{R}^{d_l \times d_h}$ and ${K}, {V} \in \mathbb{R}^{{k} \times d_h}$  represent the packed queries, keys, and values, respectively. $W_i$s are the parameters of the linear transformations,\footnote{For conciseness, all the bias terms of linear transformations in this paper are omitted.} $d_l$ is the number of query vectors, and ${d}_{k}=d_h / {n}$ is the size for each head, 
where ${n}$ is the number of heads and $d_h$ is the hidden size. The outputs of the $n$ heads are then concatenated and projected to form the final attentive representation:
\begin{align}
\mathcal{H}(Q,K,V) =
[\mathcal{A}_1;\mathcal{A}_2;\ldots;\mathcal{A}_k] {W}_{k}
\end{align}
The multihead attention is followed by a series of operations of shortcut connection, dropout, and layer normalization, which we denote as function $\mathcal{G}(\cdot,*)$, where $*$ is the input.

At each generation timestep, the input caption word $\vec{x}$ from last timestep first attends to the previously generated words $X$ to obtain the contextual information $\tilde{\vec{x}}$:
\begin{align}
    \tilde{\vec{x}} &= \mathcal{G}(\mathcal{H}_\text{x}(\vec{x}, X, X), \vec{x})
\end{align}
and then attends to the image regions $I$ and attribute words $A$ to gain the related cross-modal information $\vec{c}$:
\begin{align}
    \vec{c} &= \mathcal{G}(\mathcal{H}_\text{v}(\tilde{\vec{x}}, I, I)+\mathcal{H}_\text{a}(\tilde{\vec{x}}, A, A), \tilde{\vec{x}})
\end{align}
The parameters of $\mathcal{H}_\text{v}$ and $\mathcal{H}_\text{a}$ are shared so that their outputs are in the same space and we only use one head so that they resemble the conventional visual attention and semantic attention. The results are then transformed into $\tilde{\vec{c}}$ using a two-layer rectified linear unit $\mathcal{F}$ surrounded by $\mathcal{G}$ as well and another $\mathcal{G}$ with original input $\vec{x}$ as residual is applied to enhance the importance of the current input:
\begin{align}
\label{eq:ggf}  \tilde{\vec{c}} &= \mathcal{G}(\mathcal{G}(\mathcal{F}(\vec{c}),\vec{c}), \vec{x})
\end{align}
The vector $\tilde{\vec{c}}$ could be regarded as the local aspect vector, since it also depicts a specific aspect of the image but is not aware of the inherent relations among the individual regions or attributes. Finally, the output word is sampled from:
\begin{align}
y \sim \vec{p} = \text{softmax}({W}^{\text{C}}{\tilde{\vec{c}}})
\end{align}
where each value of $\vec{p} \in \mathbb{R}^{{|D|}}$ is a probability indicating how likely each word in vocabulary $D$ should be the current output word. In training, the whole model is trained with cross entropy loss with respect to the reference captions. The model can also be training further with reinforcement learning using a CIDEr-based reward as  \citeauthor{rennie2017self}~\shortcite{rennie2017self}.

\subsection{Global-and-Local Information Exploring-and-Distilling}

The proposed approach also takes advantage of the multi-head attention to realize the idea of learning salient region groupings and attribute collocations, which could be seen as weighted combination of the individual features.

\subsubsection{Global Visual Distilling}

When we look at an image and try to describe it, we often extend the focus on one specific object to its surrounding areas and seek for other objects that often appears together with the object. Those spatially or semantically related objects form an inherent group we attend to. Thus, visual distilling is supposed to learn region groupings that characterize the spatial or semantic relationships of each seemingly independent image regions. We use a visual self attention to achieve the effect:
\begin{align}
    \tilde{I} &= \mathcal{G}(\mathcal{H}_\text{vd}(I, I, I), I))
\end{align}
Please note that we also apply the non-linear transformation and post processing as in Eq.~(\ref{eq:ggf}), which is not included in the above equation for ease of introduction. The representation is global in that it is not coupled with specific caption contexts but learns general combinations of image regions that helps the learning of the systems. It distills naturally related image regions for a higher-level representation of the image in the vision domain and remains the same for every decoding timestep.

\subsubsection{Global Attribute Distilling}
In the language domain, we also have the ability of thinking in association and using collocations when phrasing sentences. Self-attention could also be applied to emulate the process. However, unlike image regions which are based on shapes or textures, simply combining the attributes may result in common collocations that do not actually appear in the image. The captioning system may be misled if such kind of collocations are used, which we empirically verify in the preliminary experiments. To learn meaningful collocations, we propose to use a pivot word and gather the collocations of this word, so that for each decoding timestep, a different attribute combination can be used by the decoder:
\begin{align}
    \tilde{\vec{t}} &= \mathcal{G}(\mathcal{H}_\text{ad}(\vec{x}, A, A), \vec{x})
\end{align}
As the input information is also bypassed, the collocations serve as a reference of possibly related, commonly co-occurring attributes, further lessening the risk of misinterpretation of the actual image.

\subsubsection{Global Aspect Generation}

To make use of the distilled visual and attribute knowledge and obtain a global aspect vector, the caption context is first constructed via self attention based on the input word embeddings enriched by attribute collocations:
\begin{align}
    \tilde{\vec{x}} = \mathcal{G}(\mathcal{H}_\text{x}(\tilde{\vec{t}}, \tilde{T}, \tilde{T}), \tilde{\vec{t}})
\end{align}
where $\tilde{T}$ is the pack of $\tilde{t}$. Then, the caption context is used to further incorporate the visual region groups:
\begin{align}
    \vec{c}_\text{g} = \mathcal{G}(\mathcal{H}_\text{v}(\tilde{\vec{x}}, \tilde{I}, \tilde{I}), \tilde{\vec{x}})
\end{align}
Eq.~(\ref{eq:ggf}) is used to obtain $\tilde{\vec{c}}_\text{g}$. We do not include the semantic attention such that each kind of source information is incorporated into the vector only once. The vector $\tilde{\vec{c}}_\text{g}$ could be seen as the global aspect vector, since it includes not only the regions or the attributes related to the current caption context, but also the regions and attributes that commonly show up with them. It provides a context that also explores the associative aspect of the source representations.

\subsubsection{Local Cross-Modal Distilling}

The global aspect vector gathers and distills the related cross-modal source information that is more general to the current context, which is a powerful basis for description. On the other hand, it could be too general for  word selection that is precise and detailed, since the basic unit of its sources is the learned groupings of regions and attributes. We further propose the local cross-modal distilling method to make the decoding revisit the fine-grained source information so that the exact aspect could be retrieved:
\begin{align}
    \vec{c}_\text{l} = \mathcal{G}(\mathcal{H}_\text{vl}(\tilde{\vec{c}}_\text{g}, I, I)+\mathcal{H}_\text{al}(\tilde{\vec{c}}_\text{g}, A, A), \tilde{\vec{c}}_\text{g})
\end{align}
Similar to the base modal, the parameters of $\mathcal{H}_\text{vl}$ and $\mathcal{H}_\text{al}$ are also shared. Because  $\mathcal{G}$ bypasses the global aspect vector, $\vec{c}_\text{l}$ serves as a comprehensive guide that explores and distills all of the available and presumably essential information. 
It is then fed to the output layer for word selection the same with the base model. It should be noted it is essentially the same procedure as the cross-modal attention in the base model; however, since the input is different in terms of information, it is for different purpose and functions differently. From another perspective, the base model and most of the existing models all conduct local information distilling that does not take intrinsic associations of source information into account.

The proposed approach processes the source information in such a way that from input to output, the granularity of the information goes from fine-grained to coarse-grained to fine-grained through exploring and distilling. The receptive field of the decoder is broadened at the middle of the process to accommodate the spatial and relational representation of the images and to realize an efficient flow of information.

\section{Experiment}

In this section, we describe a benchmark dataset for image captioning and some widely-used metrics, followed by our training details and evaluation of the proposed approach.\footnote{\ The code is available at \url{https://github.com/lancopku/GLIED}}

\subsection{Datasets and Metrics}

We evaluate the proposed approach on the widely-used COCO dataset \cite{chen2015microsoft}, which contains 123,287 images. Each image in the dataset is paired with 5 sentences. We use the publicly-available splits in \cite{karpathy2014deep} for offline evaluation. There are 5,000 images each in validation set and test set for COCO. 
We report results from the official COCO captioning evaluation toolkit \cite{chen2015microsoft} that uses automatic evaluation metrics SPICE \cite{anderson2016spice}, CIDEr \cite{vedantam2015cider}, BLEU \cite{papineni2002bleu}, METEOR \cite{banerjee2005meteor} and ROUGE \cite{lin2004rouge}, of which SPICE and CIDEr are specifically designed to evaluate image captioning systems. 

\subsection{Settings}

For image regions, we use the RCNN-based visual features provided by \citeauthor{anderson2018bottom}~\shortcite{anderson2018bottom}, which are extracted by Faster R-CNN. For attributes, we use the attribute prediction model pre-trained by \citeauthor{fang2015captions}~\shortcite{fang2015captions} for 1,000 attribute words on COCO. For an image, the number of attribute words is reduced to the number of image regions.

We replace caption words that occur less than 5 times in the training set with the generic unknown word token, resulting in a vocabulary with 9,487 words. The word embedding size and model size are 256 and 512, respectively, and in implementation, we share the attribute embedding and the input word embedding. 
The number of heads $n$ in multi-head attention is set to 8 unless otherwise stated. 
We train the model with both cross-entropy loss and reinforcement learning optimizing CIDEr. The model is trained with batch size of 80 for 25 epochs with early stopping based on CIDEr with cross-entropy loss, followed by reinforcement learning. 
We use Adam with a learning rate of $10^{-4}$ for parameter optimization. 
We also apply beam search with beam size $=3$ during inference. 

\begin{table}[t]

\centering
\footnotesize

\begin{tabular}{@{}l c c c c c c @{}}
\toprule
Cross-Entropy  & B-1 & B-4   & M & R   & C & S       \\ \midrule

SCST$^{\sum}$  & -         & 32.8     & 26.7     & 55.1   & 106.5 & -   \\
Up-Down  & 77.2     & 36.2  & 27.0     & 56.4    & 113.5 & 20.3 \\
RFNet$^{\sum}$ & 77.4    & 37.0 & 27.9    & 57.3    & 116.3  & 20.8 \\
GCN-LSTM   & 77.4    & 37.1 & 28.1    & 57.2    & 117.1  & 21.1 \\ \midrule
Base  &  77.0  &  36.3  & 27.6  & 56.6   &  113.5   & 20.6 \\
GLIED   & \bf 77.8    &  \bf 37.9  & \bf 28.3  & \bf 57.6   & \bf 118.2   & \bf 21.2   \\
 \midrule \midrule

RL on CIDEr & B-1 & B-4   & M & R   & C & S       \\ \midrule
SCST$^{\sum}$  & -         & 35.4     & 27.1     & 56.6    & 117.5  & -   \\
Up-Down  & 79.8     &   36.3 & 27.7     & 56.9     & 120.1  &  21.4 \\
RFNet$^{\sum}$  & 80.4     &   37.9 & 28.3     & 58.3     & 125.7  &  21.7  \\
GCN-LSTM & \bf 80.9 & 38.3 &  28.6  &  58.5  & 128.7 &  22.1 \\ \midrule
GLIED  & 80.4   & \bf  39.6& \bf28.9  & \bf58.8   & \bf  129.3  &\bf 22.6   \\
\bottomrule
\end{tabular}
\caption{Comparisons with the the existing models on the COCO Karpathy test split. 
The symbol~$^{\sum}$ denotes model ensemble.
\label{tab:res-mscoco}}
\end{table}

 \begin{figure*}[t]
 \centering
 \includegraphics[width=1.0\linewidth]{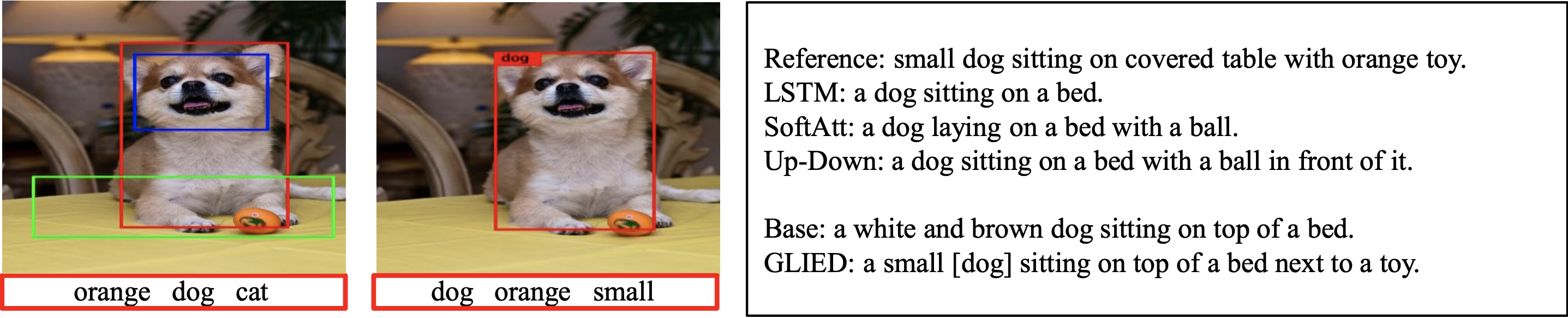}
 \caption{Examples of the generated captions and cross-modal sources. Please view in color. The first column shows the global distilling results and the second columns shows the local distilling results with image regions on the top and attribute words at the bottom. Those results are from the timestep that outputs \textit{dog} and inputs \textit{small}. For global visual distilling, we show the top-3 visual regions of the group that is most attended in local distilling. For global attribute distilling, we also show the top-3 collocations. Based on the intrinsic combinations of the source information, GLIED is able to generate more complete captions that is detailed both in the attributes and objects it describes.
 }
 \label{fig:example}
 \end{figure*}

\subsection{Experimental Results and Analysis}

In this section, we compare the proposed approach with the state-of-the-art models in terms of captioning performance and model complexity. Examples are given to show the effect of our approach. Ablation studies are conducted to verify the effectiveness of each component in the proposed approach.

\subsubsection{Quantitative Comparisons}

Four state-of-the-art models, namely SCST \cite{rennie2017self}, Up-Down \cite{anderson2018bottom}, RFNet \cite{jiang2018recurrent}, and GCN-LSTM \cite{yao2018exploring} are selected. 
The results on Karpathy test split are shown in Table~\ref{tab:res-mscoco}. For the baselines, we directly report the results from original papers. Our cross-modal base model is competitive and outperforms Up-Down, which is a strong baseline, especially in SPICE, which correlates the best with human judgment \cite{anderson2016spice}, suggesting the cross-modal point of view helps to generate coherent captions. The proposed approach (denoted as GLIED) overpasses all baselines under cross-entropy loss and reinforcement learning settings, respectively. 
In reference to Table~\ref{tab:res-subnet}, which shows sub-category scores of SPICE, GLIED does especially well in \textit{Relations} and \textit{Count}, which requires semantic and deep understanding of images, showing that the intrinsic associations of source information provides a solid basis for describing images.

\subsubsection{Qualitative Analysis}

Figure~\ref{fig:example} shows an example in comparison with the reference and the captions generated by different models. As we can see, all the models generate fluent and descriptive sentences of the image. However, they differ in how much the input information is expressed. Compared with those systems, GLIED produces a more complete and coherent caption. 

We illustrate the most attended vision region group and the top-3 most attended textual attributes in local cross-modal distilling. We also show the top-3 most attended image regions in the vision region group from global visual distilling and the top-3 most attended textual attributes in global attribute distilling.  
Global visual distilling learns to extend focus in space and seek for related regions to form a salient region group with semantics.
Global attribute distilling learns word collocations within the set of visual attributes and enriched semantic information. When the input words are descriptive (e.g., \textit{small}), it gives more attention to relevant entities (e.g., \textit{orange}, \textit{dog}, and \textit{cat}). 
The global aspect generator further gains related information in vision domain based on the enriched semantic information. Based on the collected source information (the global aspect vector), the local distilling attention retrieves original visual and textual information and decides on what is the most probable next word. For example, when inputting \textit{small}, the final attention retrieves the textual dog and the visual dog from the sources, which are much more concentrated than the global aspect vector and ensure the focus on current caption generation.

\begin{table}[t]
\footnotesize
\centering

\begin{tabular}{@{}l c c c c@{}}
\toprule
Methods                                & \#Parameters     & \specialcell{Train\\Time (h)}    & \specialcell{Inference\\Speed (ips)}   & CIDEr\\ \midrule
LSTM     & \bf 11.5M   & 16.8     & 28.6 &   105.7 \\
SoftAtt  & 12.1M   & 20.4   &  23.3   &  111.5 \\
Up-Down   & 50.1M & 24.9 &  14.8  & 113.2  \\ 
CT\ssymbol{2}  & 27.5M   &  22.7     &  12.9  & 115.1   \\ \midrule
Base  & 12.3M  &13.2 & \bf 37.9 & 113.5 \\
Ours   & 18.3M    &\bf  11.9    &   34.5 & \bf 118.2  \\ 
\bottomrule
\end{tabular}
\caption{Comparisons of model complexity and speed. \#Parameters are estimated. Time and Speed is measured on a single NVIDIA GeForce GTX 1080 Ti. ips stands for images per second. The symbol \ssymbol{2} denotes the result reported from original papers.
\label{tab:time}}
\end{table}

\begin{table*}[t]
\centering
\footnotesize
\begin{tabular}{@{}l c c c c c c c c c c c c c c c@{}}
\toprule
\multirow{2}{*}{COCO}
& \multicolumn{2}{c}{BLEU-1}
&  \multicolumn{2}{c}{BLEU-2}
&  \multicolumn{2}{c}{BLEU-3}
&  \multicolumn{2}{c}{BLEU-4}
&  \multicolumn{2}{c}{METEOR}
& \multicolumn{2}{c}{ROUGE-L}
& \multicolumn{2}{c}{CIDEr}  \\
\cmidrule(lr){2-3} \cmidrule(lr){4-5} \cmidrule(lr){6-7} \cmidrule(lr){8-9} \cmidrule(lr){10-11} \cmidrule(lr){12-13} \cmidrule(lr){14-15}

& c5 & c40 & c5 & c40 & c5 & c40 & c5 & c40 & c5 & c40 & c5 & c40 & c5 & c40 \\
\midrule
HardAtt                      &70.5           & 88.1        & 52.8     & 77.9         & 38.3  &65.8 &27.7 &53.7 &24.1 &32.2 &51.6 &65.4 &86.5 &89.3     \\
AdaAtt                             & 74.8     & 92.0     & 58.4    & 84.5     & 44.4 &74.4 &33.6 &63.7 &26.4 &35.9 &55.0 &70.5 &104.2 &105.9 \\

SCST               &78.1           & 93.7     & 61.9     & 86.0     & 47.0 &75.9 &35.2 &64.5 &27.0 &35.5 &56.3 &70.7 &114.7 &116.7     \\
LSTM-A                           & 78.7     & 93.7     & 62.7    & 86.7     & 47.6 &76.5 &35.6 &65.2   &27.0 &35.4 &56.4 &70.5 &116.0 &118.0   \\
Up-Down \  & 80.2     & \bf 95.2 &64.1     & 88.8     &  49.1 &79.4 &36.9 &68.5 &27.6 &36.7 &57.1 &72.4 &117.9 &120.5 \\
CAVP   & 80.1     &  94.9 &64.7     & 88.8     &  50.0 &79.7 &37.9 &69.0 &28.1 &37.0 &58.2 &73.1 &121.6 &123.8 \\
RFNet    & \bf 80.4     &  95.0 &\bf 64.9     & \bf 89.3     &  50.1 &80.1 &38.0 &69.2 &28.2 &37.2 &58.2 &73.1 &122.9 &125.1 \\
\midrule
GLIED    & 80.1   &  94.6 &  64.7    &  88.9   &\bf 50.2& \bf80.4 & \bf 38.5 & \bf 70.3 & \bf 28.6 & \bf 37.9 & \bf58.3 & \bf 73.8 &\bf 123.3 & \bf 125.6 \\
\bottomrule
\end{tabular}
\caption{Leaderboard performance on the online COCO evaluation server. c5 means comparing to 5 references and c40 means comparing to 40 references. 
\label{tab:res-server}}
\end{table*}

\begin{table*}[t]

\centering
\setlength{\tabcolsep}{3.5pt}

\footnotesize
\begin{tabular}{@{}l c c c c c c c c c c c c@{}}
\toprule
\multirow{2}{*}{Methods}     & \multirow{2}{*}{B-1}     & \multirow{2}{*}{B-4}     & \multirow{2}{*}{M}     & \multirow{2}{*}{R}    & \multirow{2}{*}{C}   &  \multicolumn{7}{c}{S}     \\  \cmidrule(lr){7-13} 
 & & & & &  & All &Objects &Attributes &Relations &Color &Count &Size 
 
\\ \midrule

Base  &  77.0  &  36.3  & 27.6  & 56.6   &  113.5   & 20.5 	&37.3	&10.0	&5.2	&11.6	&5.8	& 4.0 \\ 
\midrule

Base w/ Global Vis. Dist. & 77.3    & 36.5  &  27.8  & 56.9 &  115.3 &20.6	&36.9	&9.8	&5.7	&9.3	&9.2	&3.9  \\

Base w/ Global Attr. Dist.   &  77.1     &  36.9 & 27.9  & 56.8  &  114.7  & 20.8  & 36.6	&10.4	&5.5	&12.5	&  8.5	& 4.2   \\

Base w/ Global Dist. &  77.4   & 36.8  &  27.9  &  57.2  &  116.4  &  20.9	&37.3	&10.2	&5.9	& \bf 13.1	&8.7	&3.9 \\ \midrule

Base w/ Local Dist.   & 76.2        & 36.0  & 27.7  & 56.6    &  113.7  &  20.4  &  37.2      &  9.6  & 5.7   &  11.6  & 5.2    &  3.2  \\ \midrule

GLIED & \bf 77.8  &  \bf37.9  &\bf 28.3  &\bf 57.6   &  \bf 118.2   &\bf 21.2   &  \bf 38.1       & \bf 10.6  &\bf 6.2   &  11.6   &   \bf 9.4  & \bf 4.4  \\                            
\bottomrule
\end{tabular}
\caption{Results of incremental analysis of our proposed approach upon our fully-attentive cross-modal base model. 
\label{tab:res-subnet}}
\end{table*}

\subsubsection{Model Complexity and Computation Speed}

To analyze model complexity, we compare our model with (1) LSTM, which only uses a one-layer LSTM decoder \cite{vinyals2015show} but inputs visual features at each timestep as \citeauthor{lu2017knowing}~\shortcite{lu2017knowing}, (2) SoftAtt \cite{xu2015show}, which extends LSTM with visual attention, (3) Up-Down, which includes a two-layer LSTM decoder with top-down attention, (4) CT \cite{zhu2018captioning}, which is based on a transformer with six decoder blocks, and our cross-modal base model. For fair comparison, we reimplement those models using RCNN-based visual feature under the maximum log-likelihood setting. 
We also report obtained CIDEr scores. 
As Table~\ref{tab:time} shows, our model achieves arguably the best balance between speed and accuracy. Our cross-modal base model is very efficient and is comparable with Up-Down in accuracy, yet 4x smaller and 2x faster. GLIED brings about a 5-point CIDEr improvement by moderate increase in parameters with even faster training and slight inference speed regression. It suggests that the cross-modal point of view and our approach benefits deep image understanding in several aspects.

\subsubsection{Performance on the Online COCO Evaluation Server}

Following \citeauthor{jiang2018recurrent}~\shortcite{jiang2018recurrent}, we also submit our GLIED optimized using reinforcement learning to online COCO evaluation server. We compare with the top-performing entries on the leaderboard whose methods are published, which are RFNet \cite{jiang2018recurrent}, CAVP \cite{liu2018context}, Up-Down \cite{anderson2016spice}, LSTM-A \cite{yao2017boosting}, SCST \cite{rennie2017self}, AdaAtt \cite{lu2017knowing} and HardAtt \cite{xu2015show}. As we can see, the GLIED performs better than the existing systems. 

\subsubsection{Incremental Study}

We conduct a series of studies to investigate the contribution of each component in the proposed approach and the results are shown in Table~\ref{tab:res-subnet}. These experiments use cross-entropy loss. We also list the results of SPICE sub-categories to help analyze the quality and the difference of the captions. 

\paragraph{Effect of global visual distilling.}
An overall improvement is achieved when representing the visual features as region groupings. As we expected, the refined visual features are good at associating related parts in the image, which is demonstrated by the increased scores in \textit{Relations} and \textit{Count}. However, as much more information is provided, the decoder may get confused about the exact object or the attribute that is to be described, leading to impaired accuracy.

\paragraph{Effect of global attribute distilling.} 
Global attribute distilling promotes the base model in almost all sub-categories except for \textit{Object}. Compared with the self-clustered visual regions, the attribute collocations have the input word as pivot to extract constrained collocations, which are learned across examples and provide comprehensive context for details.

\paragraph{Effect of global distilling.} 
Combining the global visual distilling and the global attribute distilling gives rise to a series of multimodal representations, with correlated features being aggregated in each modality. As a result, the advantages of the region groupings and attribute collocations are united to produce a balanced improvement.

\paragraph{Effect of local distilling.} 
As we can see, incorporating local distilling directly on the base model, which is essentially a two-layer version of the base model, leads to almost the same performance, if not worse. Despite that, GLIED, which implements both global distilling and local distilling, demonstrates overall improvements. It suggests that the introduction of global combined source information induces new learning dynamics and local distilling functions differently in the new scenario. With the abundant and enriched information extracted by the global distilling method, the local distilling method helps the extraction of original and precise information, turning the cross-modal source information into further advantages in deep and semantic image understanding.

\section{Conclusions}

In this work, we present a simple yet effective approach exploring and distilling the cross-modal source information. The global distilling methods learn to capture salient region groupings and attribute collocations and explore a spatial and relational coarse-grained representation of the image, which serves as powerful basis for image descriptions. The local distilling method in contrast makes the decoder revisit the fine-grained source representation so that related and specific details can be retrieved. 
Experiments on the COCO dataset validate our proposal, which achieves 129.3 CIDEr score with fewer parameters and faster computation.

\section*{Acknowledgments}

This work was supported in part by National Natural Science Foundation of China (No. 61673028) and the Shenzhen Fundamental Research Project (No. ZDSYS201802051831427). We thank all the anonymous reviewers for their constructive comments and suggestions.

\clearpage

\bibliographystyle{ijcai19}
\bibliography{ijcai19}

\begin{thebibliography}{}

\bibitem[\protect\citeauthoryear{Anderson \bgroup \em et al.\egroup
  }{2016}]{anderson2016spice}
Peter Anderson, Basura Fernando, Mark Johnson, and Stephen Gould.
\newblock {SPICE:} {S}emantic propositional image caption evaluation.
\newblock In {\em {ECCV}}, 2016.

\bibitem[\protect\citeauthoryear{Anderson \bgroup \em et al.\egroup
  }{2018}]{anderson2018bottom}
Peter Anderson, Xiaodong He, Chris Buehler, Damien Teney, Mark Johnson, Stephen
  Gould, and Lei Zhang.
\newblock Bottom-up and top-down attention for image captioning and {VQA}.
\newblock In {\em {CVPR}}, 2018.

\bibitem[\protect\citeauthoryear{Banerjee and Lavie}{2005}]{banerjee2005meteor}
Satanjeev Banerjee and Alon Lavie.
\newblock {METEOR:} {An} automatic metric for {MT} evaluation with improved
  correlation with human judgments.
\newblock In {\em ACL}, 2005.

\bibitem[\protect\citeauthoryear{Brendel and Bethge}{2019}]{brendel2019bagnet}
Wieland Brendel and Matthias Bethge.
\newblock Approximating {CNN}s with bag-of-local-features models works
  surprisingly well on image{N}et.
\newblock In {\em ICLR}, 2019.

\bibitem[\protect\citeauthoryear{Chen \bgroup \em et al.\egroup
  }{2015}]{chen2015microsoft}
Xinlei Chen, Hao Fang, Tsung{-}Yi Lin, Ramakrishna Vedantam, Saurabh Gupta,
  Piotr Doll{\'{a}}r, and C.~Lawrence Zitnick.
\newblock Microsoft {COCO} captions: Data collection and evaluation server.
\newblock {\em arXiv preprint arXiv:1504.00325}, 2015.

\bibitem[\protect\citeauthoryear{Fang \bgroup \em et al.\egroup
  }{2015}]{fang2015captions}
Hao Fang, Saurabh Gupta, Forrest~N. Iandola, Rupesh~Kumar Srivastava, Li~Deng,
  Piotr Doll{\'{a}}r, Jianfeng Gao, Xiaodong He, Margaret Mitchell, John~C.
  Platt, C.~Lawrence Zitnick, and Geoffrey Zweig.
\newblock From captions to visual concepts and back.
\newblock In {\em {CVPR}}, 2015.

\bibitem[\protect\citeauthoryear{Geirhos \bgroup \em et al.\egroup
  }{2019}]{geirhos2019bias}
Robert Geirhos, Patricia Rubisch, Claudio Michaelis, Matthias Bethge, Felix~A.
  Wichmann, and Wieland Brendel.
\newblock Image{N}et-trained {CNN}s are biased towards texture; increasing
  shape bias improves accuracy and robustness.
\newblock In {\em ICLR}, 2019.

\bibitem[\protect\citeauthoryear{He \bgroup \em et al.\egroup
  }{2016}]{he2016deep}
Kaiming He, Xiangyu Zhang, Shaoqing Ren, and Jian Sun.
\newblock Deep residual learning for image recognition.
\newblock In {\em {CVPR}}, 2016.

\bibitem[\protect\citeauthoryear{Hochreiter and
  Schmidhuber}{1997}]{hochreiter1997long}
Sepp Hochreiter and J{\"{u}}rgen Schmidhuber.
\newblock Long short-term memory.
\newblock {\em Neural Computation}, 1997.

\bibitem[\protect\citeauthoryear{Jiang \bgroup \em et al.\egroup
  }{2018a}]{jiang2018learning}
Wenhao Jiang, Lin Ma, Xinpeng Chen, Hanwang Zhang, and Wei Liu.
\newblock Learning to guide decoding for image captioning.
\newblock In {\em {AAAI}}, 2018.

\bibitem[\protect\citeauthoryear{Jiang \bgroup \em et al.\egroup
  }{2018b}]{jiang2018recurrent}
Wenhao Jiang, Lin Ma, Yu{-}Gang Jiang, Wei Liu, and Tong Zhang.
\newblock Recurrent fusion network for image captioning.
\newblock In {\em {ECCV}}, 2018.

\bibitem[\protect\citeauthoryear{Karpathy and Li}{2015}]{karpathy2014deep}
Andrej Karpathy and Fei{-}Fei Li.
\newblock Deep visual-semantic alignments for generating image descriptions.
\newblock In {\em {CVPR}}, 2015.

\bibitem[\protect\citeauthoryear{Lin}{2004}]{lin2004rouge}
Chin-Yew Lin.
\newblock {ROUGE:} {A} package for automatic evaluation of summaries.
\newblock In {\em ACL}, 2004.

\bibitem[\protect\citeauthoryear{Liu \bgroup \em et al.\egroup
  }{2018a}]{liu2018context}
Daqing Liu, Zheng{-}Jun Zha, Hanwang Zhang, Yongdong Zhang, and Feng Wu.
\newblock Context-aware visual policy network for sequence-level image
  captioning.
\newblock In {\em {ACM} Multimedia Conference}, 2018.

\bibitem[\protect\citeauthoryear{Liu \bgroup \em et al.\egroup
  }{2018b}]{liu2018simnet}
Fenglin Liu, Xuancheng Ren, Yuanxin Liu, Houfeng Wang, and Xu~Sun.
\newblock sim{N}et: Stepwise image-topic merging network for generating
  detailed and comprehensive image captions.
\newblock In {\em {EMNLP}}, 2018.

\bibitem[\protect\citeauthoryear{Liu \bgroup \em et al.\egroup
  }{2019}]{liu2019MIA}
Fenglin Liu, Yuanxin Liu, Xuancheng Ren, Kai Lei, and Xu~Sun.
\newblock Aligning visual regions and textual concepts: Learning fine-grained
  image representations for image captioning.
\newblock {\em arXiv preprint arXiv:1905.06139}, 2019.

\bibitem[\protect\citeauthoryear{Lu \bgroup \em et al.\egroup
  }{2017}]{lu2017knowing}
Jiasen Lu, Caiming Xiong, Devi Parikh, and Richard Socher.
\newblock Knowing when to look: Adaptive attention via a visual sentinel for
  image captioning.
\newblock In {\em {CVPR}}, 2017.

\bibitem[\protect\citeauthoryear{Papineni \bgroup \em et al.\egroup
  }{2002}]{papineni2002bleu}
Kishore Papineni, Salim Roukos, Todd Ward, and Wei{-}Jing Zhu.
\newblock {BLEU}: a {M}ethod for automatic evaluation of machine translation.
\newblock In {\em {ACL}}, 2002.

\bibitem[\protect\citeauthoryear{Rennie \bgroup \em et al.\egroup
  }{2017}]{rennie2017self}
Steven~J. Rennie, Etienne Marcheret, Youssef Mroueh, Jarret Ross, and Vaibhava
  Goel.
\newblock Self-critical sequence training for image captioning.
\newblock In {\em {CVPR}}, 2017.

\bibitem[\protect\citeauthoryear{Vaswani \bgroup \em et al.\egroup
  }{2017}]{ashish2017attention}
Ashish Vaswani, Noam Shazeer, Niki Parmar, Jakob Uszkoreit, Llion Jones,
  Aidan~N. Gomez, Lukasz Kaiser, and Illia Polosukhin.
\newblock Attention is all you need.
\newblock In {\em {NIPS}}, 2017.

\bibitem[\protect\citeauthoryear{Vedantam \bgroup \em et al.\egroup
  }{2015}]{vedantam2015cider}
Ramakrishna Vedantam, C.~Lawrence Zitnick, and Devi Parikh.
\newblock Cider: Consensus-based image description evaluation.
\newblock In {\em {CVPR}}, 2015.

\bibitem[\protect\citeauthoryear{Vinyals \bgroup \em et al.\egroup
  }{2015}]{vinyals2015show}
Oriol Vinyals, Alexander Toshev, Samy Bengio, and Dumitru Erhan.
\newblock Show and tell: {A} neural image caption generator.
\newblock In {\em {CVPR}}, 2015.

\bibitem[\protect\citeauthoryear{Wu \bgroup \em et al.\egroup
  }{2016}]{wu2016what}
Qi~Wu, Chunhua Shen, Lingqiao Liu, Anthony~R. Dick, and Anton van~den Hengel.
\newblock What value do explicit high level concepts have in vision to language
  problems?
\newblock In {\em {CVPR}}, 2016.

\bibitem[\protect\citeauthoryear{Xu \bgroup \em et al.\egroup
  }{2015}]{xu2015show}
Kelvin Xu, Jimmy Ba, Ryan Kiros, Kyunghyun Cho, Aaron Courville, Ruslan
  Salakhudinov, Rich Zemel, and Yoshua Bengio.
\newblock Show, attend and tell: Neural image caption generation with visual
  attention.
\newblock In {\em {ICML}}, 2015.

\bibitem[\protect\citeauthoryear{Yao \bgroup \em et al.\egroup
  }{2017}]{yao2017boosting}
Ting Yao, Yingwei Pan, Yehao Li, Zhaofan Qiu, and Tao Mei.
\newblock Boosting image captioning with attributes.
\newblock In {\em {ICCV}}, 2017.

\bibitem[\protect\citeauthoryear{Yao \bgroup \em et al.\egroup
  }{2018}]{yao2018exploring}
Ting Yao, Yingwei Pan, Yehao Li, and Tao Mei.
\newblock Exploring visual relationship for image captioning.
\newblock In {\em {ECCV}}, 2018.

\bibitem[\protect\citeauthoryear{You \bgroup \em et al.\egroup
  }{2016}]{you2016image}
Quanzeng You, Hailin Jin, Zhaowen Wang, Chen Fang, and Jiebo Luo.
\newblock Image captioning with semantic attention.
\newblock In {\em {CVPR}}, 2016.

\bibitem[\protect\citeauthoryear{Zhang \bgroup \em et al.\egroup
  }{2006}]{zhang2006multiple}
Cha Zhang, John~C. Platt, and Paul~A. Viola.
\newblock Multiple instance boosting for object detection.
\newblock In {\em {NIPS}}, 2006.

\bibitem[\protect\citeauthoryear{Zhu \bgroup \em et al.\egroup
  }{2018}]{zhu2018captioning}
Xinxin Zhu, Lixiang Li, Jing Liu, Haipeng Peng, and Xinxin Niu.
\newblock Captioning transformer with stacked attention modules.
\newblock {\em Applied Sciences}, 2018.

\end{thebibliography}

\end{document}